\documentclass[letterpaper, 10 pt, conference]{ieeeconf}  

\IEEEoverridecommandlockouts                              

\usepackage{graphics} 
\usepackage{epsfig} 
\usepackage{mathptmx} 
\usepackage{times} 
\usepackage{amsmath} 
\usepackage{amssymb}  
\usepackage{threeparttable}
\usepackage{multirow}
\usepackage{bm}

\title{\LARGE \bf
OPA-3D: Occlusion-Aware Pixel-Wise Aggregation \\for Monocular 3D Object Detection
}

\author{{\small Yongzhi Su$^{*}$, Yan Di$^{*,\ddagger}$, Fabian Manhardt, Guangyao Zhai, Jason Rambach, Benjamin Busam, Didier Stricker, Federico Tombari}
\thanks{*Authors with equal contributions.}%
\thanks{$\ddagger$Corresponding author.}%
\thanks{$\dagger$This work was partially funded by the EU Horizon Europe Framework Program under grant agreement 101058236 (HumanTech).}
\thanks{Yongzhi Su and Didier Stricker are with 
DFKI GmbH, 67663 Kaiserslautern, Germany, and also with Faculty of Computer Science, TU Kaiserslautern, 67663 Kaiserslautern, Germany.
        {\tt\small Yongzhi.Su@dfki.de, Didier.Stricker@dfki.de}}%
\thanks{Di Yan, Guangyao Zhai, Benjamin Busam are with the Faculty of Computer Science, Technische Universit\"{a}t M\"{u}nchen, 85748 Garching bei M\"{u}nchen, Germany.
        {\tt\small shangbuhuan13@gmail.com, guangyao.zhai@tum.de, b.busam@tum.de}}%
\thanks{Fabian Manhardt is with Google, 8002 Zurich, Switzerland.
        {\tt\small fabianmanhardt@google.com}}%
\thanks{Jason Rambach is with DFKI GmbH, 67663 Kaiserslautern, Germany.
        {\tt\small Jason.Rambach@dfki.de}}%
\thanks{Federico Tombari is with the Faculty of Computer Science, Technische
Universit\"{a}t M\"{u}nchen, 85748 Garching bei M\"{u}nchen, Germany, and also with Google, 8002 Zurich, Switzerland.
        {\tt\small tombari@in.tum.de}}%
}

\begin{document}

\maketitle
\thispagestyle{empty}
\pagestyle{empty}

\begin{abstract}

Despite monocular 3D object detection having recently made a significant leap forward thanks to the use of pre-trained depth estimators for pseudo-LiDAR recovery, such two-stage methods typically suffer from overfitting and are incapable of explicitly encapsulating the geometric relation between depth and object bounding box.
To overcome this limitation, we instead propose 
to jointly estimate dense scene depth with depth-bounding box residuals and object bounding boxes, allowing a two-stream detection of 3D objects, leading to significantly more robust detections.
Thereby, the geometry stream 
combines visible depth and depth-bounding box residuals to recover the object bounding box via explicit occlusion-aware optimization. In addition, a bounding box based geometry projection scheme is employed in an effort to enhance distance perception.
The second stream, named as the Context Stream, 
directly regresses 3D object location and size.
This novel two-stream representation further enables us to enforce cross-stream consistency terms which aligns the outputs of both streams, improving the overall performance.
Extensive experiments on the public benchmark demonstrate that OPA-3D outperforms state-of-the-art methods on the main Car category, 
whilst keeping a real-time inference speed.
We plan to release all codes and trained models soon.

\end{abstract}

\section{Introduction}

Monocular 3D object detection is a very challenging, yet, essential research field in computer vision as it empowers a wide spectrum of applications, including autonomous driving~\cite{KITTI3D}, robotics manipulation~\cite{GPV-Pose}, and scene understanding~\cite{total3dunderstanding}. 
Compared to multi-sensor-based methods ~\cite{pang2020clocs,liang2019multi} that typically leverage RGB and Lidar observations jointly, monocular 3D object detection approaches~\cite{gupnet,DD3D,monocon} regress the metric 3D object center, size and heading direction from a single RGB image alone, putting forward several new challenges as an ill-posed inverse task.

The most vital challenge for the RGB-only 3D object detection is distance perception~\cite{DD3D,dis_decom,ma2021delving} as monocular 3D object detection suffers from the scale-distance ambiguity induced by the perspective projection onto the image plane. As consequence, a few pixels in image space can make large differences in 3D, which in turn makes it very hard to directly predict the pose of objects far away~\cite{ma2021delving}.

\begin{figure}[t]
    \centering
    \includegraphics[width=0.98\linewidth]{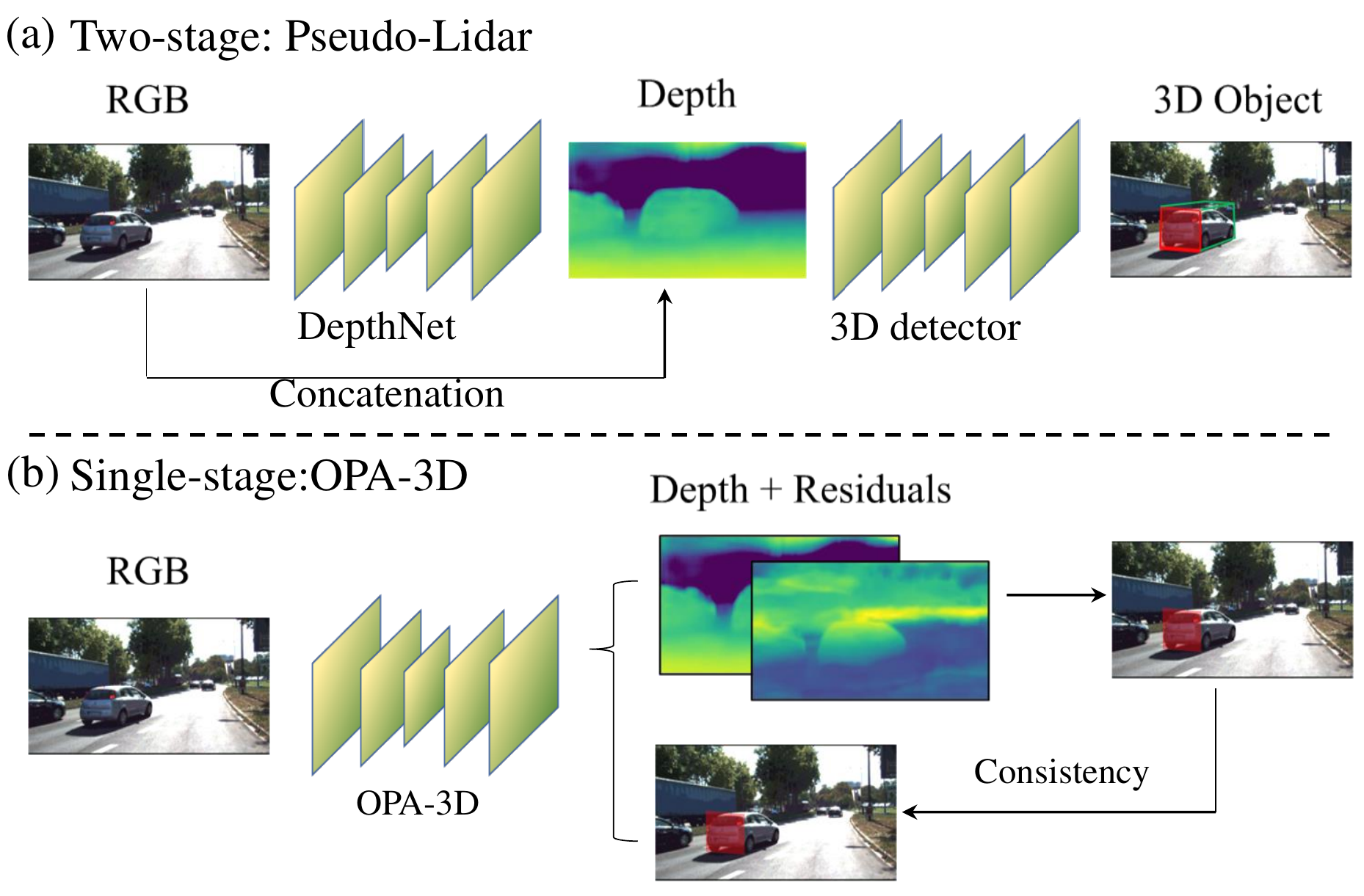}
    \caption{Compared to conventional two-stage pseudo-Lidar methods that simply append the utilized object detector to a pre-trained depth estimator, OPA-3D takes advantage of joint dense depth estimation (which can be pre-trained) and direct object bounding box regression, enabling geometry-guided cross-task consistency terms. 
    Results on public datasets demonstrate that our new network architecture outperforms all pseudo-Lidar detectors. }
    \label{fig:teasor}
\end{figure}

To overcome this limitation, pseudo-LiDAR detectors commonly employ pre-trained depth estimation networks to generate intermediate point cloud representations of the scene, in an effort to better leverage 3D scene priors. Subsequently, a 3D object detector is utilized to jointly regress object location, size and heading direction. 
The main advantage of pseudo-LiDAR methods resides in the fact that their performance on object detection can be automatically improved when updating the depth generator.
However, efficiency and generalization ability are a strong bottleneck, restricting the performance of such methods~\cite{simonelli2019disentangling}.
To address this issue and enhance distance perception at the same time, single-stage monocular 3D object detectors~\cite{gupnet,monocon,liu2021autoshape} design geometric and mathematical depth priors, e.g. localization errors~\cite{ma2021delving}, geometry projection~\cite{gupnet,dis_decom}, spatial relationship~\cite{chen2020monopair}, for the object center.
These single-stage 3D detectors demonstrate promising results on public datasets, however, their performance is still inferior to methods pre-trained with a depth dataset, with 16.46\% of Monocon~\cite{monocon} compared to 16.87\% of DD3D~\cite{DD3D} on KITTI-3D main category.

Another challenge is (self-)occlusion. In a cluttered traffic scene, objects are often occluded or truncated, which complicates inference due to insufficient context cues. Thereby, self-occlusion will remain challenging since benchmarks such as KITTI-3D only provide occlusion and truncation level and do no offer detailed ground truth such as object visibility masks. Thus, the majority of methods~\cite{DD3D,dis_decom,gupnet} simply regress all target information from the image regardless of the occlusion handling.
As exception, MonoRun~\cite{chen2021monorun} utilizes a latent vector to encapsulate occlusion, truncation and shape cues of the object.
Moreover, \cite{roddick2018orthographic} leverages view transformations to generate novel observations of the scene, mitigating the occlusion caused by perspective projection.

On the contrary, in this paper we propose an Occlusion Aware Pixel-Wise Aggregation network (OPA-3D) that takes advantage of geometry and context cues to regress 3D object bounding boxes in traffic scenes, combining the advantages of both pseudo-LiDAR and single-stage detectors. 
OPA-3D computes a shared feature embedding and subsequently stacks two streams, i.e. the Geometry Stream and the Context Stream, on top of it. As for the Context Stream, we adopt GUPNet~\cite{gupnet} to directly regress object bounding boxes from the RGB image. 
The geometry stream instead recovers the 3D information from geometrical cues. To recover the 3D bounding box, the geometry stream predicts 1) the dense depth map, and 2) the residual vectors from each point on the object surface to the corresponding bounding box faces, along with 3) the uncertainty of each residual vector. We show in Sec.~\ref{sec:BoxFromGS} that the 3D bounding box can be estimated via solving an occlusion-aware quadratic function. In addition, the geometry stream regresses the location of bounding box corners in BEV (Bird's Eye View) along the x-axis, enabling several novel 2D-3D constraints to enhance distance perception.
Thanks to this design, the dense depth map prediction can be easily pre-trained assuming the availability of a large-scale dataset, and the geometry stream provided rich geometric guidance during the training. 
Since the actual 3D bounding box can be derived from both streams, we are able to enforce consistency terms to align the individual outputs, which in turns enhances the overall performance.
In Fig.~\ref{fig:teasor}, we compare OPA-3D with conventional pseudo-LiDAR methods~\cite{wang2019pseudo,weng2019monocular}. In particular, OPA-3D is end-to-end trainable and also allows for depth pre-training.

Our main contributions can be summarized as follows:
\begin{enumerate}
\setlength{\itemsep}{0pt}
\setlength{\parsep}{0pt}
\setlength{\parskip}{0pt}
\item
We propose an \textbf{O}cclusion-Aware \textbf{P}ixel-Wise \textbf{A}ggregation network OPA-3D that encapsulates the object location and size into two information streams, i.e. the Geometry Stream and the Context Stream. The Geometry Stream provides a closed-form occlusion-aware recovery of the 3D bounding box, enabling consistency with the directly regressed object bounding box from the Context Stream, enabled both geometric and contextual guidance during the training.
\item
To enhance the distance perception, we further propose novel occlusion-aware 2D-3D geometric consistencies in BEV.

\item
OPA-3D achieves state-of-the-art performance on the real-world KITTI-3D benchmark, whilst keeping an inference speed of 25Hz, enabling potential real-time applications.
\end{enumerate}

\section{Related Work}
LiDAR-based 3D object detection methods~\cite{deng2020voxel,yin2021center} have achieved promising results in the last decade. 
In contrast, monocular 3D object detection~\cite{mousavian20173d,liu2019deep} remains an active area of research with significant space for improvement.  
In this section we highlight the three main existing branches of related work in monocular 3D object detection. 

\subsection{Monocular 3D Detection with Depth Prediction}

Taking advantage of the development of deep learning methods, especially in monocular depth estimation, approaches such as~\cite{wang2019pseudo,weng2019monocular} proposed to initially estimate the depth map from a single RGB image and subsequently transfer it into a PL (pseudo-LiDAR) point cloud. 
LiDAR-based object detectors can be directly applied to estimate the 3D object bounding box from the PL point cloud representation. 
Subsequent work attempted to strengthen this pipeline by fully leveraging the information from the RGB image~\cite{ma2019accurate,weng2019monocular}. \cite{qian2020end,ma2020rethinking} also showed improvement by training the PL methods in an end-to-end manner. Although the monocular depth estimation can be trained with rich raw video or stereo data, the task itself is still ill-conditioned and ambiguous, making it the main error source for the PL 3D object detection~\cite{wang2020task}. To overcome the limitation of imprecise distance perception, learning the depth map as an auxiliary task has been considered a better choice~\cite{DD3D} which can exploit the additional large-scale training data while not restricting the 3D object detection to the depth error. Our proposed approach also benefits from the additional available data for depth training. 

\subsection{Monocular 3D Detection with Shape Priors}
Following the work of Murthy et al.~\cite{murthy2017reconstructing}, a branch of approaches start with integrating shape prior in the 3D detection pipeline. 
Early works simplified the shape as keypoints~\cite{chabot2017deep,ansari2018earth}. Combined with CAD models, the keypoints can be utilized to determine object size or solve the object pose with a PnP (Perspective-n-Point) solver. However, the labeling of the keypoints is time-consuming and often inaccurate. To mitigate this, \cite{liu2021autoshape} automated the labeling process by proposing a deformable model-fitting pipeline. Taken into account that the understanding of object shape contributes to the object 3D detection, \cite{ku2019monocular,kundu20183d} proposed the dense object shape matching from LiDAR point cloud or using a render-based loss. Unlike explicitly applying the object shape, we encode the object shape implicitly in the depth residual prediction. We define the 3D bounding box as the optimal one given by the dense residual depth prediction, which also minimizes the influence of the inaccurate object shape annotation.

\subsection{Monocular 3D Detection with Geometry Consistency}
Deep3DBox~\cite{mousavian20173d} firstly defined the geometry consistency to lift the 2D detection to 3D detection by assuming that the 3D bounding box should fit the 2D bounding box tightly. \cite{choi2019multi} extended the prior work by formulating this geometry constraint in both image view and BEV (Bird's Eye View). However, this formulation assumes that the 2D predictions are accurate. To tackle this issue, \cite{liu2019deep,naiden2019shift,li2019gs3d} used 3D detection shifted from 2D detection only as an initial prediction. \cite{liu2019deep,ku2019monocular} generated more 3D prediction proposals based on the initial prediction, while \cite{naiden2019shift,li2019gs3d} refined the initial prediction along with other geometrically constraints, such as object orientation. 
Very recently, MonoPair~\cite{chen2020monopair} adopted the geometric relationship between objects, enabling the optimization of 3D predictions across all object instances in the image.
We propose a similar 2D-3D bounding box geometry constraint in the BEV and reduce the side effect from inaccurate 2D bounding box by adding a bounding box confidence rate.   
\section{Methodology}

Given a single RGB image, our goal is to estimate the 3 Degrees-of-Freedom (DoF) object center together with the 3 DoF object size and the 1 DoF heading direction and an optional detection confidence for each object of interest.
Moreover, in line with other works~\cite{gupnet,monocon}, we require a lot of appropriate training data for depth pre-training.
Nevertheless, as the depth prediction head of OPA-3D can be trained with LiDAR points as well as depth maps, we can easily make use of many large scale depth datasets, e.g. KITTI-Depth~\cite{KITTI3D}.
Note that only RGB images are required during inference.

\begin{figure*}[t]
    \centering
    \includegraphics[width=0.95\linewidth]{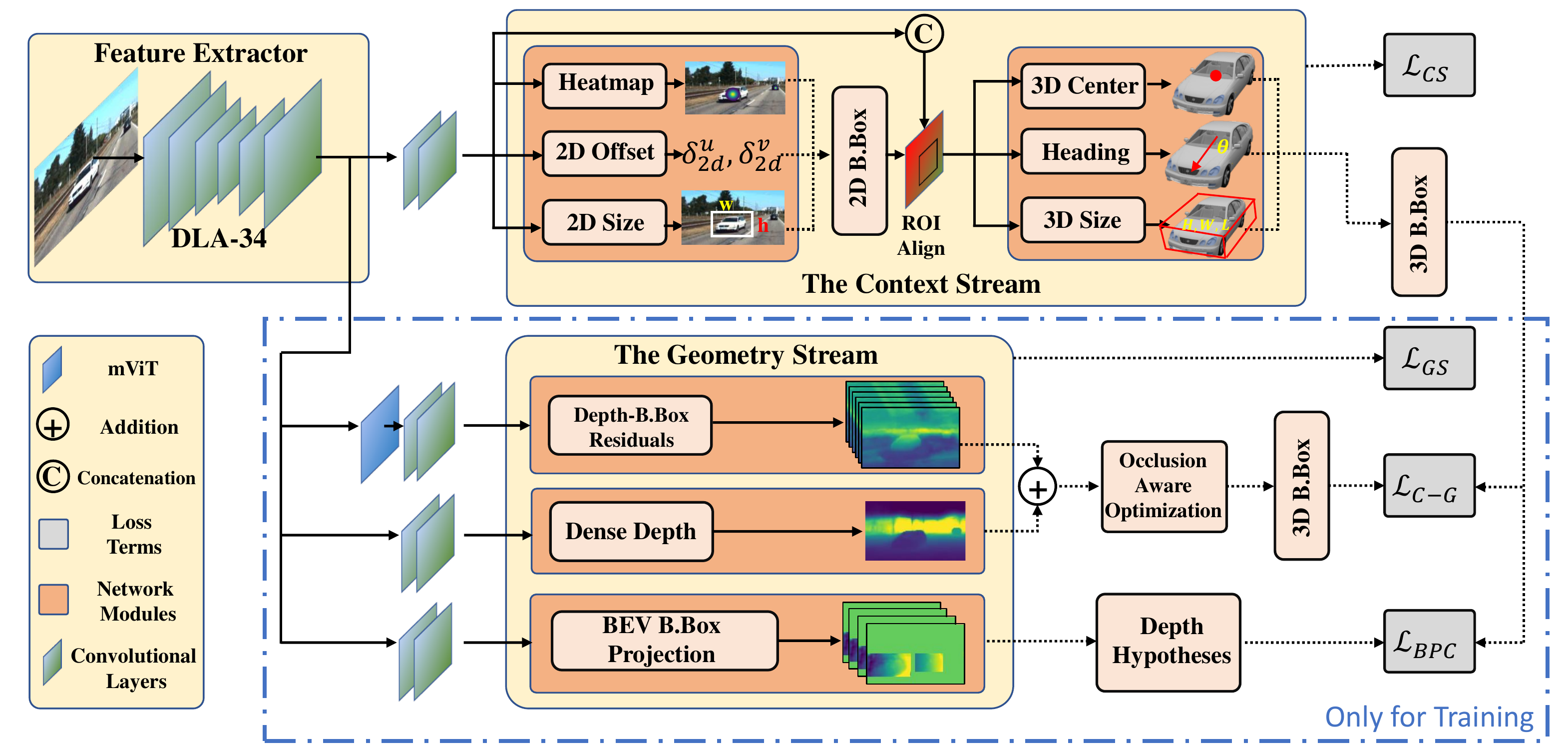}
    \caption{\textbf{Schematic overview of our OPA-3D architecture.}
    We first extract features from the given image, then the features are further processed in two streams.
    The first two heads in the Geometry Stream are used for dense depth and depth-bounding box residuals (DBR) estimation.
    By pixel-wise aggregation of the two predictions, we can recover the 3D bounding box parameters with an occlusion-aware optimization function.
    The last bounding box projection head is applied to provide 2D-3D constraints to enhance distance perception.
    We follow GUP-Net~\cite{gupnet} to design the 2D and 3D detection heads as the Context Stream.}
    \label{fig:pipeline}
\end{figure*}

\noindent \textbf{Overview. }
In contrast to pseudo-Lidar methods that detect 3D objects in a two-stage fashion, our method OPA-3D, as illustrated in Fig.~\ref{fig:teasor}, is a single-stage method which can be trained fully end-to-end.
Given a single RGB image as input, we use a backbone to extract features, which are then fed into the Geometry Stream and Context Stream to perform respective prediction tasks.  

Thereby, the first two heads in the Geometry Stream (GS) are designed for dense depth and depth-bounding box residuals (DBR) estimation. 
We show that the 3D bounding box parameters can be occlusion-aware recovered in closed-form with the pixel-wise aggregation of the output from the two heads. 
In the last bounding box projection head, we additionally predict the 2D projections of the BEV bounding box corners to enforce geometry-guided 2D-3D constraints for distance perception. 
Our Context Stream (CS) is identical to GUP-Net~\cite{gupnet}, which first predicts 2D bounding boxes and then uses ROI-Align~\cite{maskrcnn} to extract object-centered features for 3D bounding box prediction.
Compared to the CS, our GS gains more specific geometric information such as object shape and location.
We use the consistency terms ($\mathcal{L}_{C-G}$ and $\mathcal{L}_{BPC}$) to ensure the promotion of both streams and that the feature encoder can learn rich features from both streams. 
On the otherside, CS is faster and more stable since all cues are leveraged and aggregated, while GS only harnesses depth-related geometric information. Thus, we only use CS during inference.


\subsection{The Context Stream}\label{sec:ContextStream}

To localize the object in the image, we let our network predict a coarse 2D object center $\bm{C}_{o}$ in form of a heatmap, together with a 2D offset  $\bm{\delta}_{2d}=\{ \delta^{u}_{2d}, \delta^{v}_{2d} \}$ and 2D size $h, w$, so to obtain the final refined 2D bounding box using $\bm{C}_{2d}=\bm{C}_{o} + \bm{\delta}_{2d}$.
To guide the model to extract object-centered features, the 2D bounding box indicated region-of-interest (ROI) features are cropped and resized with ROI-Align~\cite{maskrcnn} and then concatenated with normalized coordinate map~\cite{roi10d} to form the input for the 3D detection head.
Finally, the 3D object center $\bm{C}^{CS}_{3d}$, heading ${\theta}$ and size $H_{CS}, W_{CS}, L_{CS}$ are predicted from the ROI features and employed to calculate the respective 3D bounding box. Notice that in the following, we use the subscript 'CS' to denote predictions from the Context Stream.
For supervision we follow the loss terms from GUP-Net~\cite{gupnet}, denoted as $\mathcal{L}_{CS}$.

\subsection{Geometry Stream}\label{sec:GeometryStream}
In the section, we introduce the three predition heads in the Geometry Stream and their supervision.

\begin{figure}[t]
    \centering
    \includegraphics[width=0.99\linewidth]{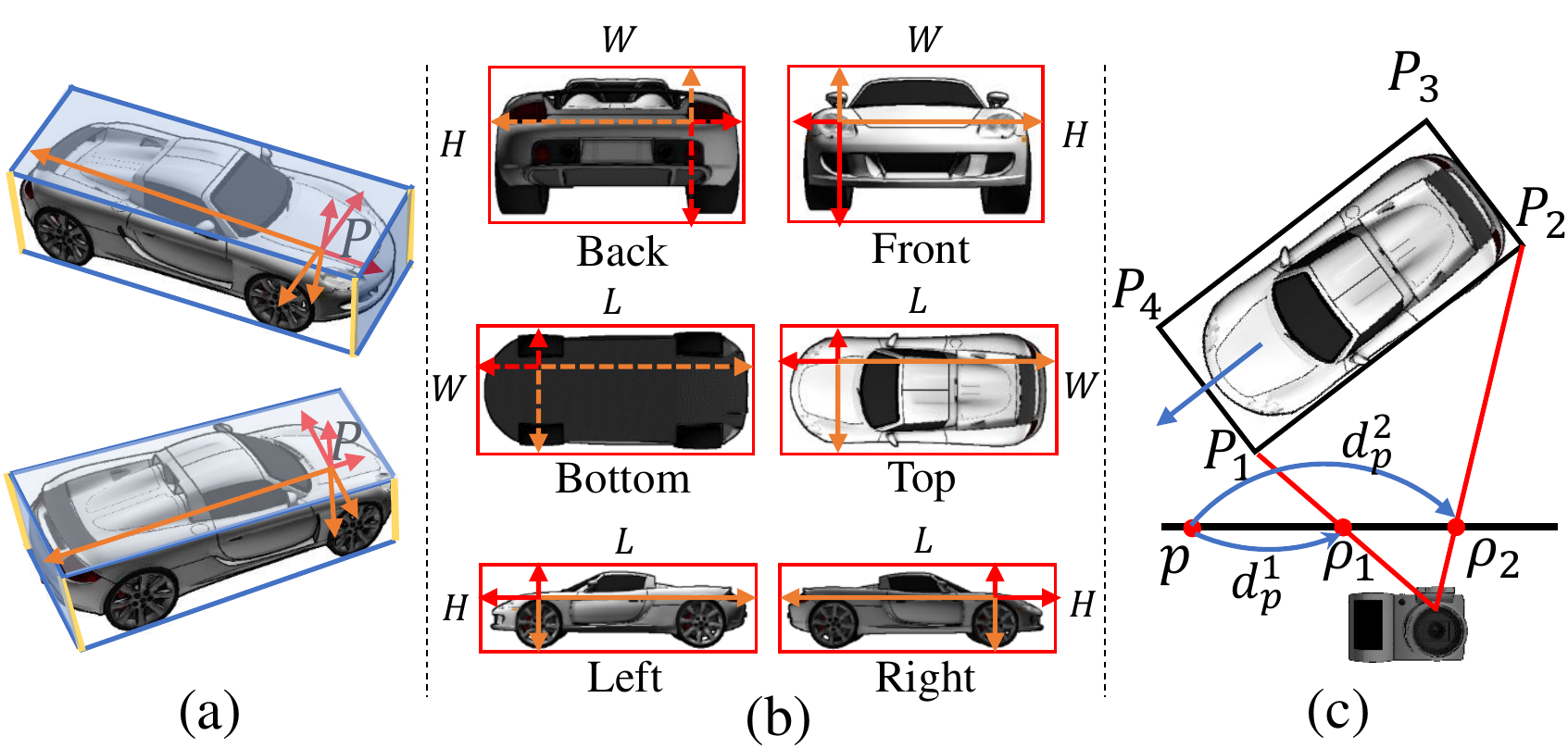}
    \caption{\textbf{Illustration of DBR and Bounding Box Projection.}
    (a) illustrates Depth-Bounding Box Residuals (DBR) in two views. Given a visible point $\bm{P}$ on the object surface, we calculate its displacement vectors $\bm{R}$ to each of the bounding box faces along the direction of the corresponding plane normal (red arrow for positive direction while orange for negative).  In (b), we demonstrate DBR from the view of the six bounding box faces.
    If $\bm{P}$ is invisible, we use dashed lines.
    $H, W, L$ refer to 3D object height, width and length.
    (c) shows the bounding box projection from the bird's-eye view.
    Since we assume the y axis points straight downwards, from the bird-eye view the surrounding lines (yellow lines in (a)) of the object box project to 4 points $\bm{P_i}, i=1,...,4$. 
    We project $\bm{P_i}$ onto the image and yield $\bm{\rho_i}, i=1,...,4$, 
    We utilize the bounding box projection head to predict the 1D displacement vector $d^{i}_{p}$ from each pixel $\bm{p}$ to the projections $\bm{\rho_i}$ along x-axis.}
    \label{fig:example_explain}
\end{figure}

\noindent \textbf{Dense Depth Estimation. } We follow Adabins~\cite{bhat2021adabins} to adopt a transformer-based network to divide the depth range into adaptive context-guided bins, which turns dense depth estimation into a linear combination of Softmax scores. The estimated depth map $\mathbf{D}$ is $1 \times H/4 \times W/4$. 

\noindent \textbf{DBR.}
DBR was first introduced in GPV-Pose~\cite{GPV-Pose} for RGB-D based category-level pose estimation. 
It describes the displacement vectors from the observed object surface towards its projections on all of the 6 bounding box faces, as shown in Fig~\ref{fig:example_explain} (a). 
We follow and develop DBR by leveraging a very small convolutional branch for regression of DBR $\mathbf{R}$ together with corresponding uncertainties $\mathbf{U}_{P}$, where $\mathbf{R},\mathbf{U}_{P} \in \mathbb{R}^{6 \times \frac{H}{4} \times \frac{W}{4}}$.

\noindent \textbf{Bird's Eye View (BEV) Bounding Box Projection. } 
The output of this head is designed for enforcing several novel 2D-3D geometric constraints for improved distance perception (details see Sec.~\ref{sec:DistanceFromBEV}).
As shown in Fig.~\ref{fig:example_explain}, for each BEV bounding box corner $\bm{P}_i$ with $i\in\{1,...,4\}$, $\bm{\rho}_i=\{\rho_i^x, \rho_i^y\}$ denotes the perspective projection of $\bm{P}_i$ onto the image plane given the known camera intrinsics. 
Since the $y$ axis is assumed to be vertical to the ground, we estimate the displacement $d^{i}_p$ and uncertainty $u^{i}_p$ from a pixel $\bm{p}$ to the location $\bm{\rho}_i$ merely along the x-axis.
The final projected locations of the corners along the x-axis can be recovered as $\rho_{i}^x = \sum_{\bm{p} \in \Omega} (p^x+d^{i}_p) * exp(u^{i}_p) / \sum_{\bm{p} \in \Omega} exp(u^{i}_p)$, where $\Omega$ includes all pixels in the 2D object bounding box and $exp(*)$ is the exponential function that transforms uncertainty scores $\bm{u_p}$ into weights.

\noindent \textbf{GS Supervision. }
Assuming the prediction follows the Laplacian distribution, one can formulate the supervision within an uncertainty framework using the Laplacian Aleatoric Uncertainty (LAU) loss~\cite{chen2020monopair}, which is defined as,

\begin{equation}
    \centering
\mathcal{L}_{LAU}(u, pr)=\frac{\sqrt{2}}{u}\left| pr-pr^{gt}\right|+log(u),
    \label{eq:LAU_Loss}
\end{equation}
where $pr$ is the prediction, $pr^{gt}$ is the corresponded ground truth and $u$ refers to the estimated uncertainty for this prediction.

Since we need to predict DBR and BEV bounding box along with their uncertainties, we supervise both prediction w.r.t the LAU loss.
We use a standard $\mathcal{L}_1$ loss to train our dense depth prediction. The final loss for the geometry stream $\mathcal{L}_{GS}$ can be computed by adding up the individual contributions of the presented three heads. 

\subsection{3D Bounding Box from Geometry Stream} \label{sec:BoxFromGS}
GPV-Pose~\cite{GPV-Pose} leverages DBR to directly establish geometry-aware consistency with the predicted pose from RGB-D images.
We extend this idea for RGB-only 3D object detection. In the following, we introduce 1) how to recover the 3D bounding box from RGB input and 2) how to solve the optimization problem in an occlusion-aware manner.

1) We explicitly recover the 3D object bounding box in closed-form via differentiable post-optimization, enabling direct supervision with the ground truth 3D object bounding box parameters for more effective feature extraction.
As shown in Fig.~\ref{fig:example_explain}, for each visible pixel $\bm{p}$, the DBR head outputs the 1D displacement vector $R^j$ that maps the back-projected 3D point $\bm{P}$ to the corresponding projection $\bm{B}_{P}^j$ on each bounding box face $j$ with $\bm{B}_{P}^j = \bm{P}+\bm{n^T_j}R^j$ and $j \in \{front, back, left, right, top, bottom \}$. 
$\bm{n_j}$ is the face normal.
We additionally estimate an uncertainty $U_{P}^j$ for each point $P$ to measure the reliability of DBR.
Since $\bm{B}_{P}^j$ is supposed to be on the bounding box face, we define the uncertainty-weighted plane fitting loss $f_s$ for all $\bm{B}_{P}^j$ as follows, 
\begin{equation}
    \centering
    \begin{split}
    f_s &= \sum_{j\in Bb_{s}}\sum _{p \in \Omega}(1-U^{j}_P)(\bm{n}_j^T\bm{B}^{j}_P-\bm{n}_j^T\bm{C}^{3d}_{GS}+\frac{S^{j}_{GS}}{2})^2\\ &+\sum_{j\in Bb_{t}}\sum _{p \in \Omega}(1-U^{j}_P)(\bm{B}^j_P - \bm{C}^{3d}_{GS} + \frac{S^{j}_{GS}}{2})^2 ,
    \end{split}
    \label{fs}
\end{equation}
Thereby, $Bb_s$ contains the $\{front, back, left, right\}$ faces of the 3D bounding box, and $Bb_t$ contains the $\{top, bottom\}$ faces.
$H_{GS}, W_{GS}, L_{GS}, \bm{C}^{3d}_{GS}$ are the target bounding box height, width, length and the 3D center needed to be recovered from the GS.
$S^j_{GS} \in \{-L_{GS}, -W_{GS}, L_{GS}, W_{GS}, H_{GS}, -H_{GS}\}$ for corresponding $j \in \{Bb_s \cup Bb_t\}$, e.g. $S_{GS}^{back} = -W_{GS}$ in case of $j=back$.  
Since the target parameters $H_{GS}, W_{GS}, L_{GS}, \bm{C}^{3d}_{GS}$ are decoupled in Eq.~\ref{fs}, we can calculate the partial derivatives of $f_s$ w.r.t. these variables $\{\frac{\partial f}{\partial W_{GS}}, \frac{\partial f}{\partial L_{GS}}, \frac{\partial f}{\partial H_{GS}}, \frac{\partial f}{\partial \bm{C}^{3d}_{GS}}\}$ and directly recover $H_{GS}, W_{GS}, L_{GS}, \bm{C}^{3d}_{GS}$ by setting the partial derivatives to be zero and solving the equations. 

2) Trying to minimize Eq.~\ref{fs} alone could lead $U_{P}^j$ converge to $1.0$. However, as a confidence term, $U_{P}^j$ should have a geometrically meaning, such as reflecting the occlusions. Therefore, we introduce the regularization term in the form of
\begin{equation}
\centering
    \begin{split}
    f_t=\alpha  \sum _{p,j}(U^{j}_P) (W_{GS}-W_{\xi})^2 + \beta \sum _{p,j}(U^{j}_P)(L_{GS}-L_{\xi})^2 \\
    + \gamma \sum _{p,j}(U^{j}_P)(H_{GS} - H_{\xi})^2,
    \label{ft}
    \end{split}
\end{equation}

where $W_{\xi}, L_{\xi}, H_{\xi}$ are prior mean 3D object size.
$\alpha, \beta, \gamma$ are small constant weights to incorporate prior size knowledge for regularization.
The regularization term with $\{U_P^j, \alpha, \beta, \gamma\}$ reflects our occlusion handling strategy.
For the occluded bounding box faces in $j$, the corresponding uncertainty $U_P^j$ is large, signalizing that it is hard to predict their location directly.
As shown in Fig.~\ref{fig:example_explain} (b), when considering the \textit{front} view, the \textit{back} side is completely self-occluded by the object. Thus, without any prior knowledge, it is infeasible to infer the object's length from this view.
It is worth noting that this also applies to occlusion caused by other objects.
Therefore, forcing the network to estimate these ill-posed values leads to confusion and, hence, converges in performance deterioration.
We solve the optimization problem objective to $f = f_s + f_t$ to mitigate this problem.
If $U_P^j$ for a certain bounding box face $j$ is very large (or in other words the prediction from the image is not reliable), the corresponding term in $f_s$ will be small. In this case, the 3D bounding box will be predicted close to the prior size $\{W_{\xi}, L_{\xi}, H_{\xi}\}$ to ensure the minimum of $f_t$.

To summarize, our optimization-based bounding box recovery explicitly describes the pipeline from image to depth and finally to the object bounding box.
Since $H_{GS}$, $W_{GS}$, $L_{GS}$, $\bm{C}^{3d}_{GS}$ are solved in a differentiable manner, we can directly supervise the Geometry Stream with ground truth bounding box parameters besides the respective supervision on dense depth and DBR, enforcing the network to learn more effective features for 3D object detection.

\subsection{Object Distance from BEV Bounding Box Projection}\label{sec:DistanceFromBEV}
To enhance distance perception, geometric projection of object height is widely harnessed in literature~\cite{gupnet,ma2021delving}. In this paper, we further extend this idea and propose object width and length related geometric priors to establish several novel 2D-3D constraints.
As shown in Fig.~\ref{fig:example_explain}, we have
\begin{equation}
    \centering
    \left\{\begin{matrix}
z_1\rho_1^x=\mathbf{K}\bm{P}_1=\mathbf{K}(\bm{C}+\frac{W}{2}\bm{n}_2+\frac{L}{2}\bm{n}_1), \\[6pt]
z_2\rho_2^x=\mathbf{K}\bm{P}_2=\mathbf{K}(\bm{C}+\frac{W}{2}\bm{n}_2-\frac{L}{2}\bm{n}_1),
\end{matrix}\right.
\end{equation}
where $z_1, z_2$ are the $z$ coordinates of $P_1$ and $P_2$, $K$ denotes the camera intrinsic matrix and $W, L, C$ are the target bounding box parameters.
When solving the above equation, we can obtain the depth $z_C$ of the object center as follows
\begin{equation}
    \centering
    z_C =\frac{f_xn_x+c_xn_z}{\rho_1^x-\rho_2^x}L-\frac{n_z(\rho_1^x+\rho_2^x)}{2(\rho_1^x-\rho_2^x)}L-\frac{n_x}{2}W, \label{eq:zc}
\end{equation}
where $f_x$, $c_x$ are camera focal length and its optical center along the $x$ axis.
Further, $\bm{n}_1=[n_x, 0, n_z]^T$ is the normalized heading direction of the object (\textit{c.f.} blue arrow in Fig.~\ref{fig:example_explain} (c)).
Note that we again only utilize the $x$ coordinate of the BEV bounding box projections.
Since the $y$ axis points straight downwards, the projections of the surrounding lines (yellow lines in Fig.~\ref{fig:example_explain} (a)) are parallel to the $y$ axis on the image plane, thus regressing their $x$ coordinate via pixel-wise aggregation is more reliable than exactly locating their 2D positions. We can build a 2D-3D constraint based on each BEV bounding box edge, resulting in up to 4 constraints. 

\subsection{Geometry Consistency Loss}\label{sec:consistency_loss}
In this section, we introduce how to build cross-stream consistency losses between the predictions from the geometry stream and the context stream. 

\noindent \textbf{Two-Stream Consistency. } Notice that our proposed model is able to predict the 3D object bounding box from both presented streams. Whereas the geometry stream recovers the 3D bounding box via optimization of Eq.~\ref{fs} and Eq.~\ref{ft}, which is fully differentiable, the context stream instead directly regresses all target parameters $\{H_{CS}, W_{CS}, L_{CS}, \bm{C}^{{3d}}_{CS}\}$ from the RGB image.
As consequence, we can naturally enforce consistency between both streams according to
\begin{equation}
    \centering
    \begin{split}
    \scalebox{0.75}{
    $\mathcal{L}_{C-G} = \left\| H_{GS} - H_{CS}\right\| + \left\| W_{GS} - W_{CS}\right\| + \left\| L_{GS} - L_{CS}\right\| + \left\| \bm{C}^{GS}_{3d} - \bm{C}^{CS}_{3d}\right\|.$
    }
    \end{split}
\end{equation}

\noindent \textbf{Bounding Box Projection Consistency. }
For each object we can generate up to 4 hypotheses for depth $z_C$ following Eq~\ref{eq:zc}. Thus, given the predicted 3D length and width from the Context Stream, we define our consistency loss for object depth as
\begin{equation}
    \centering
    \mathcal{L}_{BPC} = \sum_{j\in BL}\omega_j\left\| z^j_c - C^{z}_{CS}\right\|,
    \label{lbpcd}
\end{equation}
where $C^{z}_{CS}$ is the $z$ coordinate of $\bm{C}^{3d}_{CS}$, $BL$ contains all four corner pairs of BEV bounding box edges and
$\omega_j$ measures the weight of each term.
When considering $\bm{\rho}_1$ and $\bm{\rho}_2$, $\omega_{12}$ is defined as $\omega_{12}=v*[1-exp(-k|\rho_1^x-\rho_2^x|)]$, where $k$ is a constant defined according to the heading direction, $v=1$ if $\{\bm{P}_1, \bm{P}_2\}$ are visible and otherwise $v=0$.
This weight definition enforces that a visible projection with larger observation angle is preferred.

\subsection{Overall Loss Function}

The overall loss function is a combination of all four previously introduced loss terms, 
\begin{equation}
    \mathcal{L}_{Overall} = \mathcal{L}_{CS} + \mathcal{L}_{GS} + \mathcal{L}_{C-G} + \mathcal{L}_{BPC}.
\end{equation}
Thereby, the loss terms will be automatically weighted with Hierarchical Task Learning (HTL) scheme as proposed in GUP-Net~\cite{gupnet}, which allows to define preliminary tasks for a specific loss. This specific loss will then start to be used when its preliminary task has been properly trained. 
The ranking of tasks in the Context Stream are identical to GUP-Net~\cite{gupnet}. 
The Geometry Stream starts with dense depth prediction and BEV bounding box prediction, which in turn have no preliminary tasks. 
In contrast, DBR prediction requires dense depth estimation as preliminary task. The two consistency terms kick in once the predictions become stable.
In essence, both consistency losses depend on the entire Context Stream. 
Moreover, the $\mathcal{L}_{C-G}$ also requires stable dense depth computation and DBR prediction, while the $\mathcal{L}_{BPC}$ relies on a steady BEV bounding box prediction.

\section{Experiments}
In this section we first introduce our experimental setup and provide our utilized implementation details, before presenting our results for the \textbf{KITTI-3D}~\cite{KITTI3D} dataset. 

\subsection{Experimental Setup}
\noindent \textbf{Datasets. }
We provide detailed evaluation on the commonly-used KITTI-3D~\cite{KITTI3D} benchmark dataset.
We use the official KITTI-3D split of 7481 training and 7581 testing images.
We further follow~\cite{gupnet,dis_decom,DD3D} and split the training data into a training and validation subset, each containing around 3.7k samples, for ablation purposes.
During training we also leverage the provided LiDAR points, calibration files and ground truth bounding box annotations. In this paper, we mainly focus on three common categories, i.e. \textit{car, pedestrian, cyclist}.
Following previous works~\cite{gupnet,monocon}, we conduct ablation studies mainly on the \textit{car} category.

\noindent \textbf{Implementation Details. }
For a fair comparison with the state-of-the-art, we use DLA-34~\cite{dla34} with a downsampling ratio of 4 as our backbone network for feature extraction. We train the network with a batchsize of 8 using the AdamW~\cite{loshchilov2018fixing} optimizer. The learning rate increased from $1e-5$ to $1.25e-3$ in the first 5 epochs using a cosine warmup. We train our networks for 200 epochs and decay the learning rate by 0.1 after the 110-th and 150-th epoch. We empirically set $\alpha, \beta, \gamma$ in Sec.~\ref{sec:BoxFromGS} to $1e-3$ for occlusion handling. 
The input images have been resized to $1280\times380$. We augment the input data by randomly cropping and flipping of the input images, as well as random brightness changes. The network has been pretrained with the KITTI-Depth dataset~\cite{KITTI3D} by enabling only the dense depth prediction head for 25 epochs using the clean training split (removed the overlapping of the test images in KITTI-3D)~\cite{DD3D}.

\noindent \textbf{Evaluation Metrics. }
Following the evaluation protocol of most current works, we report the AP (Average Precision) for the IoU metric at a 0.7 threshold for 3D detection~\cite{monocon,gupnet,DD3D}. We approximate the area underneath the Area-Precision curve using 40 points as proposed in \cite{simonelli2019disentangling} . 

\subsection{Comparison with state-of-the-art}
\noindent \textbf{Results for Car on KITTI-3D. }
We first compare with other state-of-the-art (SOTA) monocular 3D detectors on the KITTI-3D \textit{test} set for the \textit{car} category in Tab.~\ref{tab:kitti_test_car}. 
It can be seen that OPV-3D exceeds all other SOTA methods and shows an obvious improvement ($2.03\%$) over the GUP-Net~\cite{gupnet} baseline on the moderate criteria. Similarly, we are also on par or better than Monocon~\cite{monocon}, which is similar to GUP-Net~\cite{gupnet}, yet leverages several auxiliary tasks to improve 3D detection. In contrast to the auxiliary tasks in Monocon~\cite{monocon}, our proposed geometry stream is able to utilize the depth information for training, thus achieving larger performance improvements.

Note that DD3D~\cite{DD3D} leverages an in-house dataset with 15M images for depth pre-training, together with a feature pyramid network based on the much larger V2-99 backbone~\cite{v299}. Nonetheless, despite the use of this large amount of training data for depth, we can still slightly exceed DD3D on \textit{test} and, especially, on \textit{validation} with an 3D IoU of 19.40\% \textit{vs.} 16.92\% with the same backbone.  

\begin{table*}[t]
\centering

\caption{{Comparison with state-of-the-art methods of the \textit{car} category on KITTI-3D.}
}
\begin{threeparttable}
\scalebox{0.95}{
\begin{tabular}{c|c|ccc|ccc|ccc|ccc|c}
\hline
\multirow{3}{*}{Method}  & \multirow{3}{*}{Extra Data}  & \multicolumn{6}{c|}{ Test Split } &  \multicolumn{6}{c|}{ Validation Split } & \multirow{3}{*}{\shortstack{Time\\(sec)}}\\
\cline{3-14}
 & & \multicolumn{3}{c|}{ $AP_{3D|R40|IOU\geq0.7}$ } & \multicolumn{3}{c|}{$AP_{BEV|R40|IOU\geq0.7}$ } & \multicolumn{3}{c|}{ $AP_{3D|R40|IOU\geq0.7}$ } & \multicolumn{3}{c|}{ $AP_{BEV|R40|IOU\geq0.7}$ } & \\
& & Mod. & Easy & Hard & Mod. & Easy & Hard & Mod. & Easy & Hard & Mod. & Easy & Hard &  \\
\hline
ROI-10D~\cite{roi10d} & None & 2.02 & 4.32 & 1.46 & 4.91 & 9.78 & 3.74 & 6.63 & 9.61 & 6.29  & 9.91& 14.50& 8.73& 0.2 \\
MonoPair~\cite{chen2020monopair} & None & 9.99 & 13.04 & 8.65  &14.83 &19.28 &12.89 & 12.30 & 16.28 & 10.42  &18.17 &24.12 &15.76 & 0.057 \\
DLE~\cite{ma2021delving} & None & 12.26 & 17.23 & 10.29  & 18.89 & 24.79 & 16.00 & 13.66 & 17.45 & 11.68 & 19.33 & 24.97 & 17.01 & 0.040 \\
GUP-Net~\cite{gupnet} & None & 15.02 & 22.26 & 13.12 & 21.19 & 30.29 & 18.20 & 16.46 & 22.76 &	13.72  & 22.94 & 31.07 & 19.75 & 0.034 \\
Monocon~\cite{monocon} & None & 16.46 & 22.50 &	13.95  & 22.10 & 31.12 &19.00 & 19.01 & \textbf{26.33} & 15.98  & - & - & - & 0.026 \\
\hline
DD3D~\cite{DD3D} & Depth$^{*}$ & \underline{16.87} & \underline{23.19} & \textbf{14.36}  &\textbf{23.41} &32.35 &\textbf{20.42} & (16.92) & - & -  & (24.77) & - & - & - \\ 
\hline
D4LCN~\cite{D4LCN} & Depth & 11.72 & 16.65 & 9.51  & 16.02 & 22.51 & 12.55 & - &	- & -  & -  & -&- & 0.20 \\
CaDNN~\cite{cadnn} & LiDAR & 13.41 & 19.17  & 11.46  & 18.91 & 27.94 & 17.19 & 16.31 & 23.57 & 13.84 & - & - & - & 0.63 \\ 
AutoShape~\cite{liu2021autoshape} & CAD Model & 14.17 & 22.47 & 11.36  &20.08 & 30.66& 15.59& 14.65 & 20.09 & 12.07 & - &- &- & 0.050 \\
MonoDistill~\cite{chong2022monodistill} & LiDAR & 16.03 & 22.97 & 13.60 & \underline{22.59} & 31.87 & \underline{19.72} & 18.47 & 24.31 & 15.76 & \underline{25.40} & \underline{33.09} & \textbf{22.16} & 0.040 \\
Ours & Depth & \textbf{17.05} & \textbf{24.60} & \underline{14.25}  & 22.53 & \textbf{33.54} & 19.22 & \textbf{19.40} & \underline{24.97} & \textbf{16.59} & \textbf{25.51} & \textbf{33.80} & \underline{22.13} & 0.040 \\

\hline
\end{tabular}
}
\footnotesize
Overall best results are in bold and the second best results are underlined.  
Methods are ranked according to the moderate AP for the IoU metric at a 0.7 threshold as given in the KITTI-3D benchmark.
Note that on the test split DD3D utilizes deeper V2-99~\cite{v299} instead of DLA-34 as the feature extractor, and also pre-trains the network on the unreleased large-scale DD3D15M dataset, while OPA-3D only uses DLA-34 and pretrains on the clean split of KITTI-Depth dataset~\cite{DD3D} for fair comparison with the SOTA. 
On the validation split, we report the result of DD3D with DLA-34~\cite{dla34} as the backbone.

\end{threeparttable}

\label{tab:kitti_test_car}
\end{table*}
\begin{table}
\centering
\caption{{Comparison with state-of-the-art methods of the \textit{pedestrian} and \textit{cyclist} on KITTI-3D \textit{test} split. $AP_{3D|R40|IOU\geq0.7}$ is reported.}
}
\begin{threeparttable}

\begin{tabular}{c|ccc|ccc}
\hline
\multirow{2}{*}{Method}  & \multicolumn{3}{c|}{ Pedestrian } &  \multicolumn{3}{c}{ Cyclist } \\
\cline{2-7}
& Mod. & Easy & Hard & Mod. & Easy & Hard \\
\hline
MonoPair~\cite{chen2020monopair}  & 6.68 & 10.02 & 5.53 & 2.12 & 3.79 & 1.83  \\
DLE~\cite{ma2021delving}  & 6.55 & 9.64 & 5.44 & 2.66 & 4.59 & 2.45  \\
GUP-Net~\cite{gupnet}  & 9.76 & 14.95 & 8.41 & \underline{3.21} & 5.58 &	2.66  \\
Monocon~\cite{monocon}  & 8.41 & 13.10 &	6.94 & 1.92 & 2.80 & 1.55  \\
\hline
DD3D$^{*}$~\cite{DD3D}  & \textbf{11.04}& \textbf{16.64} & \textbf{9.38} & \textbf{4.79} &  \textbf{7.52} & \textbf{4.22}  \\ 
\hline
D4LCN~\cite{D4LCN}  & 3.42 & 4.55 & 2.83 & 1.67 & 2.45 & 1.36  \\
CaDNN~\cite{cadnn}  & 8.14 & 12.87  & 6.76 & 3.41 & \underline{7.00} & \underline{3.30}  \\ 
Ours  & \underline{10.49} & \underline{15.65} & \underline{8.80} & 3.45 & 5.16 & 2.86  \\

\hline
\end{tabular}
\footnotesize
Overall best results are in bold and the second best results are underlined. 
\end{threeparttable}

\label{tab:kitti_test_other}
\end{table}
\noindent \textbf{Results for Pedestrian and Cyclist on KITTI-3D. }
Interestingly, we notice larger improvements over Monocon~\cite{monocon} for the \textit{Pedestrian} and the \textit{Cyclist} categories, hinting that its auxiliary tasks cannot be learned well with small objects. However, we also observed that OPV-3D is less effective on the \textit{Cyclist} category. We attribute this drop in performance to the thin structure of cyclists. As our model is trained with depth from multiple Lidar point clouds, missing depth information for cyclists induces a weakened perception of depth.

\subsection{Ablation Study}
In Tab.~\ref{tab:kitti_ablation} we present several ablation studies on the KITTI-3D \textit{validation} split w.r.t. the \textit{Car} category. Row 'A1' refers to the results of the baseline model~\cite{gupnet}. In row 'A2', we research the improvement introduced by learning dense depth prediction as an auxiliary task. Although we observed a large improvement, we show in the following that the learning tasks in our proposed Geometry Stream contribute more.   

\noindent \textbf{Effect of Two-Stream Consistency Loss. } In this experiment, we disable the BEV bounding box head and $\mathcal{L}_{BPC}$. The GS recovers 3D bounding box from dense depth and DBR prediction, and the $\mathcal{L}_{C-G}$ ensures the consistency preditions from GS and CS. When removing this branch, we notice a drop on \textit{Car} under moderate criteria (Tab.~\ref{tab:kitti_ablation} B1 vs A0, also B1 vs A1), proving the effectiveness of our proposed $\mathcal{L}_{C-G}$. Since the $\mathcal{L}_{C-G}$ requires dense depth map prediction, we are able to easily pre-train the network on a depth dataset. After pre-training with the KITTI-Depth dataset~\cite{KITTI3D}, the results further increased by $0.61\%$. This improvement indicates that the training of $\mathcal{L}_{C-G}$ benefits from an additional depth dataset, which can be transformed from Lidar raw data without annotation effort.

\noindent \textbf{Effect of BEV bounding box projection. } We also want to ablate the usefulness of $\mathcal{L}_{BPC}$, by removing the dense depth prediction, DBR prediction and $\mathcal{L}_{C-G}$. The model again outperforms the baseline model clearly by $2.24\%$ (Tab.~\ref{tab:kitti_ablation} C vs A1), which supports the effectiveness of $\mathcal{L}_{BPC}$.   
As no depth information is used, the improvements by $\mathcal{L}_{BPC}$ are overall slightly less high than by $\mathcal{L}_{C-G}$. 

\noindent \textbf{Effect of combination of both consistency loss. } Finally, by integrating both consistency losses into the baseline model, we can obtain the results for our proposed OPV-3D network. OPV-3D exceeds the models having only one consistency term.
In summary, taking advantage of the additional depth training data and our geometric consistencies, OPV-3D is capable of yielding SOTA results.

\begin{table}[t]
\centering
\caption{{Ablation study of the \textit{car} category on KITTI-3D \textit{val} split.}
}
\begin{threeparttable}

\begin{tabular}{c|l|ccc}
\hline
\multirow{2}{*}{Stage}  & \multirow{2}{*}{ Geometry Stream } & \multicolumn{3}{c}{3D $AP_{IoU \geq 0.7}$ } \\
\cline{3-5}
& & Mod. & Easy & Hard \\
\hline
A1 & Baseline (GUP-Net) & 15.76 & 22.01 & 13.11 \\
A2 & A1 + Dense Depth Prediction & 17.31 & 22.76 & 14.58  \\
\hline
B1 & A2 + $\mathcal{L}_{C-G}$ & 18.10 & 24.81 & 15.27  \\
B2 & B1 + pretrained depth prediction & 18.77 & 25.81 & 15.65  \\
\hline
C & A1 + $\mathcal{L}_{BPC}$ & 18.00 & 24.51 & 15.15 \\
\hline
D & B2 + C  & 19.40 & 24.97 & 16.59  \\

\hline
\end{tabular}
\footnotesize
\end{threeparttable}

\label{tab:kitti_ablation}
\end{table}

\subsection{Qualitative Results}
\begin{figure}[t]
    \centering
    \includegraphics[width=0.99\linewidth]{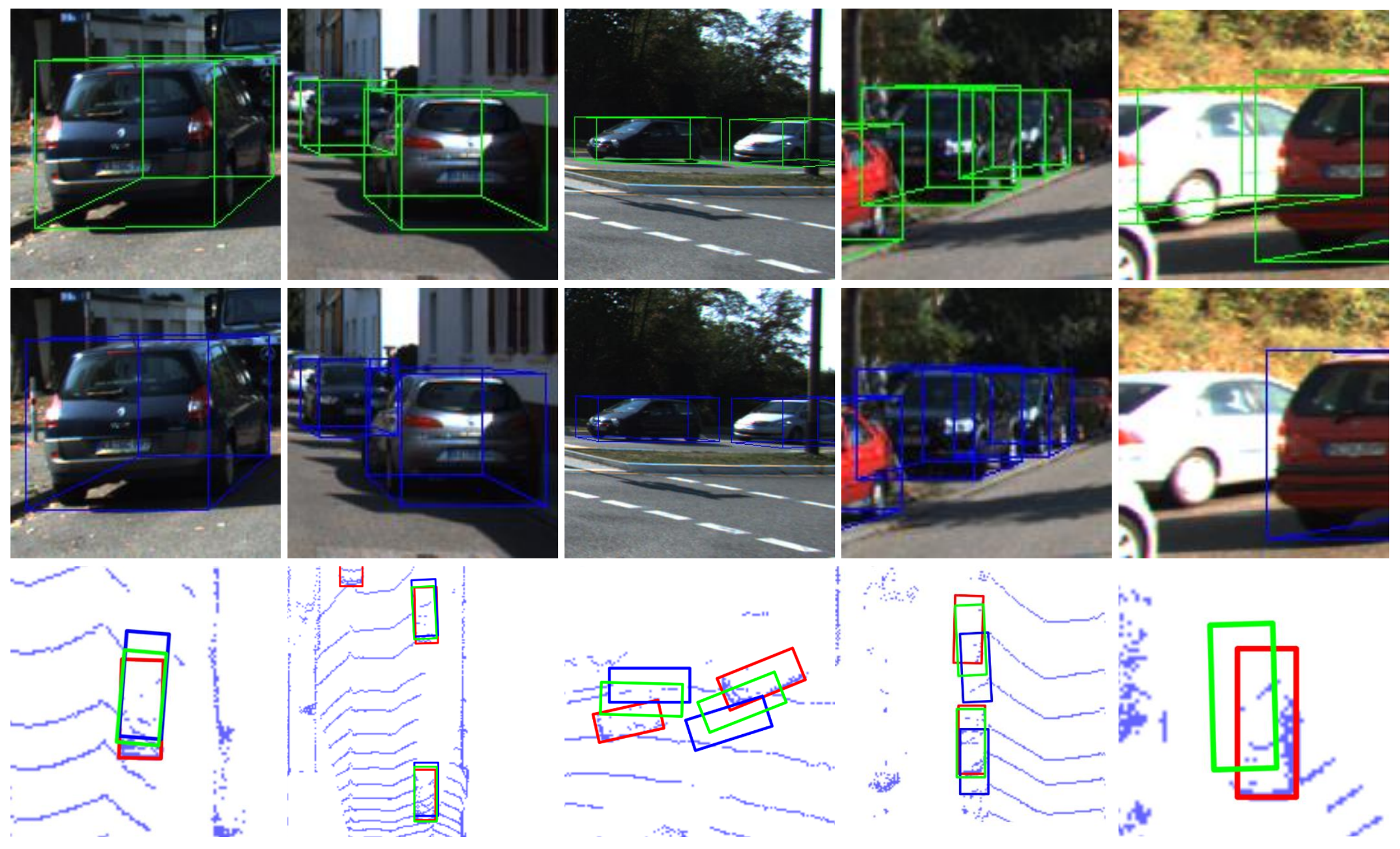}
    \caption{We visualize example predictions on the KITTI-3D validation set. Our OPA-3D predictions are visualized in green, while the prediction from the baseline model~\cite{gupnet} are shown in blue. We also render the ground truth BEV bounding box in red. We notice that OPV-3D produces more accurate predictions for the bounding box faces under self-occlusion as well as external occlusions.} 
    \label{fig:QualitativeResults}
\end{figure}

We show some qualitative results in Fig.~\ref{fig:QualitativeResults}. OPA-3D predictions are visualized in green, while the prediction from the baseline model~\cite{gupnet} are shown in blue. We also render the ground truth BEV bounding box in red. Further, in the first three columns, the car's front, back, or right side is self-occluded, while in the last two column, the car is occluded by other objects. As can be observed, OPA-3D is able to predict more accurate bounding boxes under (self-)occlusion.

\section{Conclusion}
In this paper 
we proposed a novel Occlusion-Aware Pixel-Wise Aggregation network OPA-3D, which jointly predicts the object bounding box from the geometry stream and the context stream, allowing for cross-branch consistency.
We further propose to leverage the bounding box projection to establish several 2D-3D constraints to promote distance prediction.
Extensive experiments on public benchmarks demonstrate that OPA-3D achieves state-of-the-art performance, whilst being able to achieve real time performance.

\bibliographystyle{IEEEtran}
\bibliography{egbib}

\end{document}